\documentclass{article}

\PassOptionsToPackage{numbers, compress}{natbib}

\usepackage[main, final]{neurips_2026}

\usepackage[utf8]{inputenc} 
\usepackage[T1]{fontenc}    
\usepackage{hyperref}       
\usepackage{url}            
\usepackage{booktabs}       
\usepackage{amsfonts}       
\usepackage{nicefrac}       
\usepackage{microtype}      
\usepackage{xcolor}         
\usepackage{graphicx} 
\usepackage{amsmath} 
\usepackage{colortbl}
\usepackage{multirow}
\usepackage{hhline}
\usepackage{bm}

\usepackage[most]{tcolorbox}
\usepackage{enumitem}
\usepackage{caption}
\usepackage{ bbold }
\usepackage[most]{tcolorbox}

\usepackage{wrapfig}

\usepackage{booktabs,xcolor}
\definecolor{barbg}{HTML}{DDF1FA}
\definecolor{barfg}{HTML}{B3E3F7}
\newlength\barw 
\newcommand\barmin{0.40}\newcommand\barmax{0.55}
\newcommand\bc[1]{%
  \makebox[0pt][l]{\color{barbg}\rule[-.6ex]{\barw}{2.5ex}}%
  \makebox[0pt][l]{\color{barfg}\rule[-.6ex]{\fpeval{(0#1-\barmin)/(\barmax-\barmin)}\barw}{2.5ex}}%
  \makebox[\barw][r]{#1\,}}

\newtcolorbox{takeaway}{
  colback=black!3,
  colframe=black,
  boxrule=0.9pt,
  sharp corners,
  left=5pt,
  right=5pt,
  top=3pt,
  bottom=3pt
}

\newif\ifshowtrims
\showtrimstrue 

\title{%
What's Holding Back Latent Visual Reasoning?
}

\author{
André G. Viveiros$^{1,2,3}$\thanks{Corresponding author: andre.viveiros@lisboa.tecnico.pt} \quad
Nuno Gonçalves$^{1,2,4}$ \quad
André F. T. Martins$^{1,2,3}$ \quad
Matthias Lindemann$^{2}$
\\[0.5em]
$^{1}$ Instituto Superior Técnico, Universidade de Lisboa \quad
$^{2}$ Instituto de Telecomunicações \\
$^{3}$ Gandara \quad
$^{4}$ Carnegie Mellon University
}

\begin{document}

\maketitle

\begin{abstract}
    
Humans can approach complex visual problems by mentally simulating intermediate visual steps, rather than reasoning through language alone. Inspired by this, several works on Vision-Language Models have recently explored chain-of-thought reasoning with continuous latent tokens as intermediate visual ``imagination'' steps. 
In this work, we investigate how recent models leverage such latent tokens. Surprisingly, we find that model accuracy is unaffected when latent tokens are replaced by uninformative ``dummy'' tokens.
This indicates that latent tokens play a minimal causal role in the model's final prediction.
To better understand this phenomenon, we analyze both the training signal provided by oracle latent representations and the quality of the latent tokens generated at inference time. Our experiments reveal two crucial issues holding back latent visual reasoning: 
First, in most existing datasets, oracle latent tokens provide limited additional information beyond the original image and do not substantially simplify the task, leading models to ignore them during training and effectively \emph{bypassing} them at inference time. 
When fine-tuned on a diagnostic dataset, in which latent tokens provide sufficient support for the final prediction, we show that models can causally rely on them.
Second, the latent tokens produced at inference time deviate from their corresponding oracle representations, \emph{collapsing} to a narrow region and preventing benefits even when the model relies on them.
Overall, our findings suggest that future progress in latent visual reasoning depends on two key pillars: high-quality datasets with informative intermediate steps and more precise latent token prediction. The code, models, and datasets are \href{https://github.com/GuilhermeViveiros/LanteRn.git}{publicly available}.

\end{abstract}

\section{Introduction}

Vision-Language Models (VLMs) have achieved strong performance across a wide range of visual tasks \citep{bai2025qwen3vltechnicalreport, 
wang2025internvl35advancingopensourcemultimodal,
vteam2026glm45vglm41vthinkingversatilemultimodal}. However, they continue to struggle with scenarios that require spatial and compositional reasoning 
where solving a problem depends on the internal construction and manipulation of visual representations rather than purely textual descriptions \citep{tang2025legopuzzlesgoodmllmsmultistep,xu2025visulogicbenchmarkevaluatingvisual, fu2024blinkmultimodallargelanguage}. 
In such cases, relying solely on text-based decoding can limit the model’s ability to process complex visual relationships.

\begin{figure}
    \centering
    \includegraphics[width=0.95\linewidth]{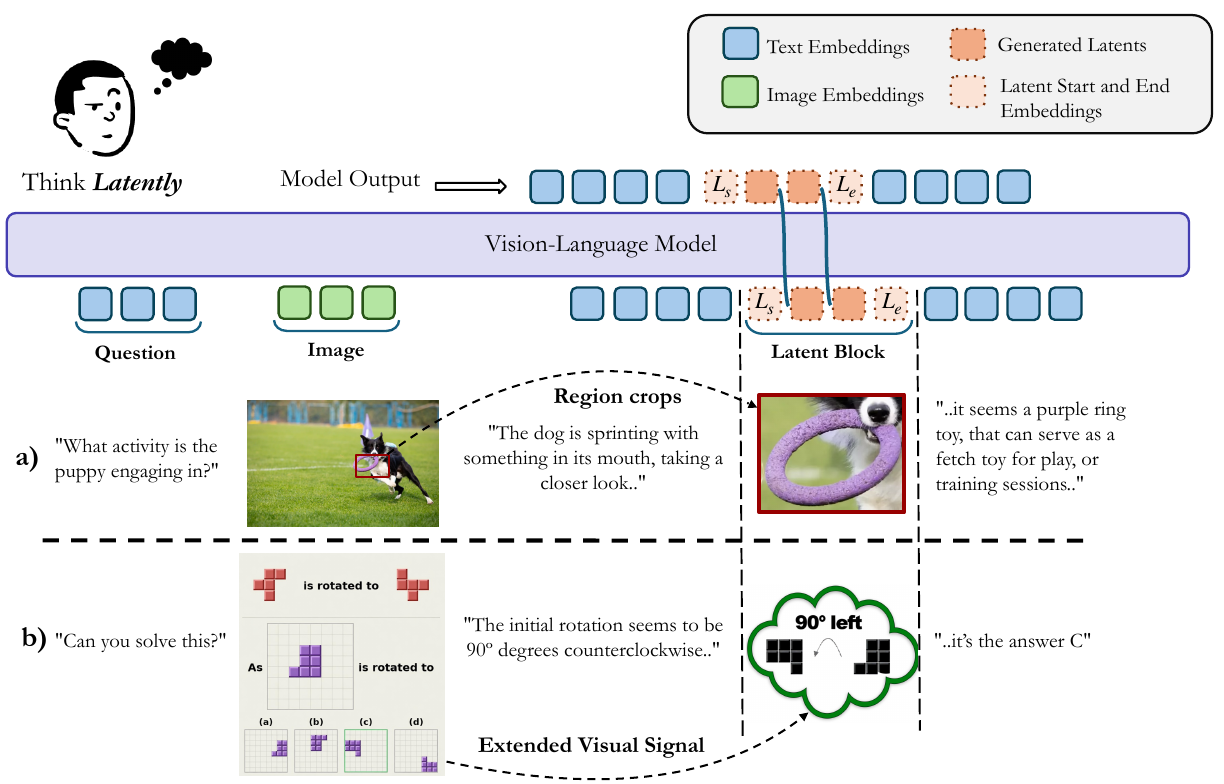}
    \caption{The usefulness of latent tokens depends on their content. \textit{a)} When they encode a subregion of the input, it provides little auxiliary signal, resulting in the model ignoring them. \textit{b)} When they capture task-relevant non-trivial processing of the input (e.g., spatial relations), this provides a stronger incentive for the model to integrate latent tokens into its reasoning to improve its answer.}
    \label{fig:overview}
\end{figure}

Latent visual reasoning has recently been explored as a pathway to address this limitation \citep{wang2025monetreasoninglatentvisual, dong2026interleavedlatentvisualreasoning}. These approaches introduce continuous latent tokens as a dedicated space for \emph{visual} chain-of-thought (CoT) reasoning, allowing models to move beyond text-based reasoning to manipulate visual representations, analogous to how humans approach complex visual problems \citep{shepard1971mental, kosslyn1996image}. Similarly to textual CoT, visual CoT decomposes a complex problem into simpler steps, but the steps operate on visual representations, such as focusing on a specific part of an image or transforming objects (see Figure~\ref{fig:overview}).
In principle, such latent tokens could enable models to visually ``imagine'' task-relevant transformations and provide a way for more structured and spatially aware reasoning.

Prior work has reported performance improvements by using latent visual reasoning, suggesting that it can improve model capabilities \citep{li2025latentvisualreasoning, wang2025monetreasoninglatentvisual, viveiros2026lanternlatentvisualstructured}. 
Yet, it remains unclear what these latent representations actually encode, and to what extent they causally influence model predictions.

In this work, we investigate the role of latent tokens in visual reasoning tasks by evaluating their causal impact on the model's answer by replacing latent tokens with severe interventions. We find that these interventions have little to no effect on model accuracy, and in some cases, performance even slightly improves when latent tokens are completely removed. In addition, accuracy barely changes when models are conditioned on oracle latents at inference time, despite these latents containing task-relevant intermediate visual information.
We see these observations as symptoms of two failure modes in current VLM training setups, a \textit{latent bypass problem} and \textit{latent representation collapse}.

The \textit{latent bypass problem} refers to the tendency of the model to ignore latent representations at inference time. We show that this issue arises from the data used in current VLM training setups (see Figure \ref{fig:overview}a), where latent tokens rarely provide additional information that is helpful for predicting the correct answer, reducing the incentive for the model to integrate them into its reasoning process.
Fine-tuning experiments on a diagnostic dataset involving analogical reasoning over Tetris-like shape rotations support this hypothesis and we observe that models do rely on latent tokens when they provide information that would otherwise not be readily available in the original input.

The \textit{latent representation collapse} refers to the tendency of generated latents to converge toward highly similar and weakly informative representations, limiting their usefulness for reasoning at inference time. Across several models, we observe that generated latent representations exhibit extremely high similarity across samples but comparatively lower similarity to corresponding ground-truth latents, suggesting that the models fail to construct diverse and discriminative visual abstractions. 
Our main contributions are as follows:
\begin{enumerate}[leftmargin=!]
	\item We show that four recent off-the-shelf latent visual reasoning models \citep{li2025latentvisualreasoning,wang2025monetreasoninglatentvisual,dong2026interleavedlatentvisualreasoning,viveiros2026lanternlatentvisualstructured} largely ignore the latent tokens in producing their answers (\textit{latent bypass problem}).
	\item We identify a major factor for the latent bypass problem: training data with latent tokens that only provide easily extractable information, such as image subregions.
    In the context of rotating Tetris-like shapes, we show that a model can learn to rely on latent tokens if they encode a non-trivial transformation of the input image.
    
	\item Finally, we show that current methods predict latents that collapse to a narrow region of the latent space but tend to be relatively far from the ground truth (\textit{latent representation collapse}).
\end{enumerate}



\section{
Background: Visual Reasoning in Latent Space
}
\label{latent_comparison}
Latent visual reasoning methods differ in how latent representations are derived and supervised during training. To provide the context needed for our analyses, we outline the training recipes of four recent models that cover representative design choices in latent supervision: \emph{\textbf{LVR}} \citep{li2025latentvisualreasoning}, \emph{\textbf{Monet}} \citep{wang2025monetreasoninglatentvisual}, \emph{\textbf{ILVR}} \citep{dong2026interleavedlatentvisualreasoning} and \emph{\textbf{LanteRn}} \citep{viveiros2026lanternlatentvisualstructured}.

At a high level, these models augment standard autoregressive decoding with latent reasoning segments, delimited by special tokens such as \texttt{latent\_start} ($L_s$) and \texttt{latent\_end} ($L_e$) as illustrated in Figure~\ref{fig:overview}. Upon generating an $L_s$ token, the model transitions from generating discrete tokens to producing continuous latent representations, using its hidden states without projecting them through the language modeling head. 
These latent states are generated autoregressively, until a latent block of fixed length is complete (e.g., 6 to 12 steps, a tiny fraction of the token budget consumed by visual inputs).
Once the final step is reached, an $L_e$ token is automatically inserted, and the model resumes standard text generation conditioned on both the previous text inputs and latent representations.

\begin{figure}
    \centering
    \includegraphics[width=0.95\linewidth]{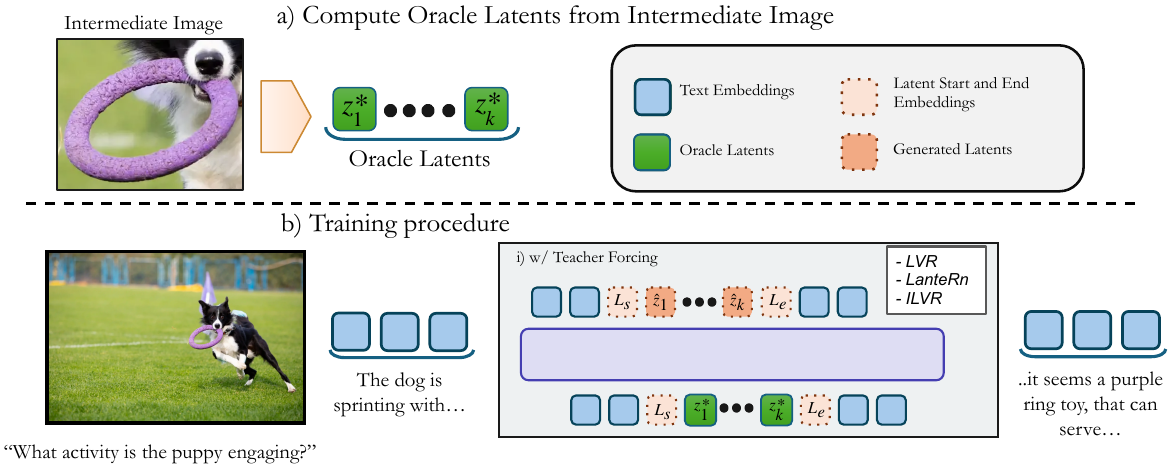}
    \caption{General framework. \textit{a)} Oracle latent tokens are computed from intermediate visual representations. \textit{b)} Training is extended into the continuous latent space, where the model predicts latent tokens, mainly conditioned on oracle latent tokens via teacher forcing (i.e., $\bm{z}^*_1,\dots$, $\bm{z}^*_k$).
 }
    \label{fig:training_pipe}
\end{figure}

During training, the model is optimized using a combination of the standard cross-entropy objective for next-token prediction and a latent alignment objective that encourages the generated latent sequence $\hat{\bm{z}}$ to remain close to the oracle latent sequence $\bm{z}^{*}$. The overall training objective is given by

\begin{align}
    \mathcal{L} = \underbrace{-\frac{1}{T}\sum_{t=1}^{T} \log P_{\theta}(y_t \mid y_{<t}, \bm{z}, x)}_{\text{Next Token Prediction}} + \gamma \underbrace{\frac{1}{K} \sum_{k=1}^{K}
\ell(\hat{\bm{z}}_k, \bm{z}^*_k)}_{\text{Latent Alignment}}.
\label{eq:latent_loss}
\end{align}
Here, $\gamma$ controls the contribution of the latent alignment objective, $K$ denotes the number of latent tokens, and $\ell$ represents either the mean squared error (MSE) or cosine distance between the latent and oracle representations. 
If a model is trained with teacher forcing, then $\bm{z}$ is instantiated as $\bm{z} := \bm{z}^*$, otherwise $\bm{z} := \hat{\bm{z}}$.
All approaches considered here
follow a shared training paradigm where 1) the model is first trained to align its generated latents $\hat{\bm{z}}$  to the target oracle latents $\bm{z}^*$ derived from auxiliary visual inputs, often referred to as \emph{intermediate images} (see Figure \ref{fig:training_pipe}a).
Subsequently, the latent alignment loss is removed and 2) training shifts towards optimizing downstream task performance, typically through RL, where $\bm{z} := \hat{\bm{z}}$ is used as intermediate auxiliary representations rather than reconstruction targets. 
We focus primarily on the
first
stage, because it is responsible for introducing visual latent reasoning capabilities, making it central to our analysis. Additionally, as we will show in Section~\ref{section3}, models exhibit similar behavior in both stages.

\textbf{LVR} and \textbf{LanteRn} \citep{li2025latentvisualreasoning, viveiros2026lanternlatentvisualstructured} 
use a two-stage training pipeline, where the model is first trained with oracle latents provided as input context to align its generated latent tokens $\hat{\bm{z}}$ with oracle latents $\bm{z}^*$ (see Figure~\ref{fig:training_pipe}b). These oracles are derived from embeddings computed from intermediate images, corresponding to cropped subregions of the original input. The embeddings are obtained by passing each intermediate image through a frozen vision encoder, followed by average pooling to reduce the resulting features into a fixed set of $K$ latent tokens.
In the second stage, the latent alignment objective is removed and RL is used to optimize task performance, where supervision is applied only on text tokens.
The two methods differ in how oracle latents are constructed: \emph{LVR} computes embeddings directly from the full-image and selects the embeddings corresponding to the target subregion, preserving global context, whereas \emph{LanteRn} encodes intermediate images independently, allowing more flexibility in handling diverse intermediate inputs.

\textbf{ILVR} \cite{dong2026interleavedlatentvisualreasoning} extends prior latent reasoning by generating multiple blocks of $K$ latent tokens rather than relying on a single latent block. The alignment loss is applied to all latent blocks. To provide stable supervision, ILVR uses a momentum teacher model, implemented as an exponential moving average of the online parameters, which generates targets conditioned on the ongoing reasoning process and intermediate image.

\textbf{Monet} \cite{wang2025monetreasoninglatentvisual} refines the construction of oracle latents while also using cropped subregions as intermediate signal. It first fine-tunes the model with interleaved text-image inputs to encourage intermediate visual information. It then uses the resulting model as a teacher, distilling latent supervision from its hidden states conditioned on these intermediate images. In subsequent stages, intermediate images are removed, allowing the model to internalize the learned latent structure. Unlike prior methods, Monet does not condition on oracle latents $\bm{z}^*$. Instead, it feeds the generated latent token $\hat{\bm{z}}_t$ as input for the next step $t+1$, effectively training in a free-running manner (see Figure~\ref{fig:overview}).

\textbf{Training Data:} 
In most existing work, latent supervision is grounded in cropped subregions, or visual rationales that preserve the underlying content with minimal modification (e.g, masking an object).
\textbf{ILVR} uses \emph{COMT} \citep{cheng2025comt} and \emph{VSP} \citep{wu2024vspassessingdualchallenges} as primary data sources. These datasets contain visual rationales derived from the original figure, preserving its underlying content while guiding the model through a more perceptually grounded reasoning process.
\textbf{LanteRn}, \textbf{LVR} and \textbf{Monet} all use \emph{VisCoT} \citep{shao2024visual} as the primary data source. \emph{VisCoT} is annotated with intermediate bounding boxes highlighting key subregions essential for answering the questions (see Figure~\ref{fig:training_pipe} for an example). 
The choice of alignment data directly shapes what visual information the model learns to encode in latent space, and is therefore central to our analysis.
We show in the subsequent sections that these data sources are insufficient to incentivize models to meaningfully rely on latent tokens to improve their predictions.


\section{Latent Tokens Have Little to No Causal Effect on Reasoning}
\label{section3}
We investigate the causal role of latent tokens across four latent visual reasoning architectures,  \emph{LanteRn}, \emph{LVR}, \emph{Monet} and \emph{ILVR} (described in Section \ref{latent_comparison}). We take off-the-shelf checkpoints and run each of them as is (standard inference). That is, we autoregressively predict the latent tokens $\hat{\bm{z}}$ and produce the answer via greedy decoding from $P_{\theta}(y | \bm{z}=\hat{\bm{z}}, \bm{x})$. To understand the causal role of latent tokens in the model's decision, we then intervene on the latent tokens and replace $\hat{\bm{z}}$ by some continuous tokens $\bm{z}'$, effectively decoding from $P_{\theta}(y | \text{do}(\bm{z}=\bm{z}'), \bm{x})$ in the notation of \citep{pearl1994probabilistic}.
If there is an intervention where $\bm{z}'$ contains no information about the correct answer and yet the model continues to predict the correct answer, this indicates the model does not rely on $\bm{z}$ for its final prediction.
We consider the following interventions to replace the latent tokens $\hat{\bm{z}}$ with uninformative tokens $\bm{z}'$:
\begin{description}[leftmargin=0pt]
	\item[Random Subregion:] selecting a cropped subregion of an \textit{unrelated} image and constructing oracle latents as used during training, injecting plausible but task-irrelevant information (see Appendix \ref{app:subregion-extraction}).
    \item[Zeros:] using a constant value, $\bm{z}' = [L_s, \mathbf{0}_{1:K}, L_e]$.
	\item[Noise:] sampling Gaussian noise, $\bm{z}' = [L_s, \epsilon_{1:K}, L_e], \quad \epsilon_k \sim \mathcal{N}(0,1)$.
	\item[Skip Latents:] disabling the latent pathway by inserting \texttt{latent\_end} ($L_e$) after \texttt{latent\_start} ($L_s$), i.e.\ setting $\bm{z}'=[L_s, L_e]$, forcing the model to resume text-only generation. This also removes the extra computation steps, ruling out a pause-token-like compute benefit \citep{goyal2024think}.
\end{description}

The first intervention based on random subregions produces latent tokens that are in-distribution. Crucially, they provide no additional signal beyond the question and the input image. Hence, for a model that relies on the latent tokens, we expect this intervention to push the model towards different predictions, resulting in a large drop in accuracy. 
The remaining interventions are out-of-distribution and contain no information about the correct answer at all.

We evaluate how models react to the interventions on two benchmarks, BLINK and $V^*$Bench. BLINK \citep{fu2024blinkmultimodallargelanguage} is a perception-heavy benchmark, where we focus on tasks involving object localization and spatial reasoning. $V^*$Bench \citep{vstar} evaluates fine-grained visual detail search and relative spatial reasoning. These benchmarks were chosen to probe capabilities closely aligned with the training data of the evaluated models, avoiding scenarios without supervision for latent reasoning. In Appendix~\ref{app:more-benchmarks}, we extend the analysis to GQA \citep{hudson2019gqa} and BabyVision \citep{chen2026babyvision}, which require more explicit multi-step reasoning, and find that overall conclusions are very similar.

\begin{figure}
	\centering
	\includegraphics[width=0.5\linewidth]{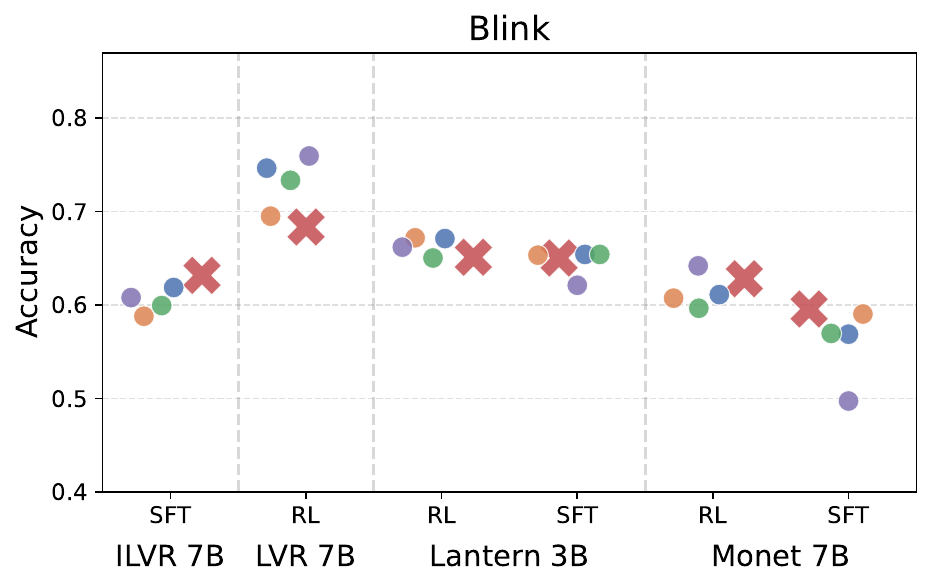}%
	\includegraphics[width=0.485\linewidth]{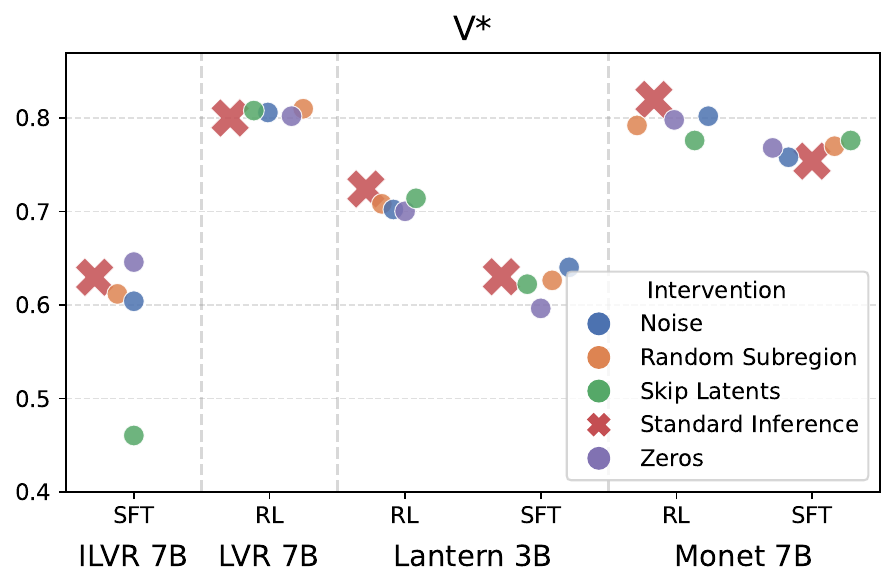}%
	\caption{Accuracy when performing standard inference vs interventions on latents on Blink and V$^*$. Replacing latents generated by the model with non-informative dummy tokens largely maintains performance. For every model and dataset, there is always at least one intervention that maintains accuracy within 2 percentage points, demonstrating a limited effect of latents on the models' decisions.}
    
	\label{fig:interventions}
\end{figure}

As shown in Figure \ref{fig:interventions}, we observe a consistent pattern across all models: performance remains largely unchanged under interventions to latent tokens, and in some cases even slightly exceeds standard inference (red crosses). This behavior is also consistent across both the SFT stage and after RL. Some out-of-distribution interventions are too harsh for some models, such as skipping latents for ILVR on V$^*$. However, \textbf{there is always at least one intervention} that replaces the latent tokens with uninformative tokens and \textbf{maintains roughly the same accuracy}. Note that finding a single intervention with comparable accuracy already demonstrates that the latent tokens generated with the standard inference procedure are not necessary to achieve that level of performance.

\setlength{\tabcolsep}{5pt}
\begin{table}[t]
\begin{minipage}{.6\linewidth}
    \small
	\centering
	\caption{Accuracy on \textbf{VisCoT} when replacing latent tokens. Despite access to ground-truth, models show only marginal gains over standard inference and the interventions. 
    }
    \vspace{0.2cm}
	\label{tab:id_evals}

    \begin{tabular}{lllllll}
    \toprule
                   & \textbf{ILVR} & \textbf{LVR}    & \multicolumn{2}{c}{\textbf{LanteRn}} & \multicolumn{2}{c}{\textbf{Monet}} \\
                   \cmidrule(lr){2-2} \cmidrule(lr){3-3} \cmidrule(lr){4-5} \cmidrule(lr){6-7}
                    & & & RL           & SFT          & RL          & SFT         \\
                   \midrule
Random Subregion   & \textbf{64}    & 73             & 82        & 78             & 81        & 75        \\
Noise              & 61             & 74             & 82         & \textbf{80}    & 82        & 71        \\
Zeros              & 59             & 76             & 82         & 79 & 85        & 41        \\
Skip Latents       & 60             & \textbf{78}    & 82         & 79 & 83        & \textbf{78}        \\[1.5ex]
Standard Inference & 62             & 73             & \textbf{83}         & 78                & 83        & 76        \\
\textit{Oracle Latents}& 63 & 77 & \textbf{83}         & 79    & \textbf{86}        & 77   \\

\bottomrule
\end{tabular}
\end{minipage}
\hspace{0.2cm}
\begin{minipage}{.38\linewidth}
    \vspace{0.13cm}
	\centering
	\small
	\caption{Accuracy on \textbf{VisCoT} with \textit{LanteRn}.
		\textbf{Std}: train/eval on filtered original data.
		\textbf{Masked}: relevant region in image is masked;
		\textbf{Pause Tokens}: model trained with dummy tokens.
	}
	\label{tab:masking}
	\vspace{0.22cm}
	\begin{tabular}{lrr}
		\toprule
		& \textbf{Std.}
		& \textbf{Masked}  \\
		\midrule
		Oracle latents & 79 & \textbf{75}   \\
		Random Subregion& 78 & 61 \\
		Noise& 79 & 59   \\
		Zeros& \textbf{80} & 59  \\
		\midrule
		Pause Tokens  &  79 & 59 \\
		\bottomrule
	\end{tabular}
\end{minipage}
\end{table}

A possible explanation for these observations is that standard inference produces low-quality latent tokens that models cannot effectively use, while they \textit{could} benefit from higher-quality ones.
To test this, we turn to VisCoT \citep{shao2024visual}, which \textit{LVR}, \textit{LanteRn} and \textit{Monet} use as a primary source of training data. In VisCoT, we have access to ground-truth intermediate images, that contain relevant information to solve the task. If the models genuinely rely on the latent pathway for reasoning, conditioning on these oracle representations should yield a measurable improvement in performance.



As shown in Table~\ref{tab:id_evals}, providing oracle latent tokens does not lead to consistent performance improvements, with results remaining largely unchanged across models. Significance tests support this: on BLINK, $V^*$ and VisCoT, no model performs significantly better with its own latents than under every intervention (Appendix~\ref{app:significance}).
In combination with our findings that models are also insensitive to interventions to latent tokens that remove task-relevant information, this highlights a fundamental limitation: 

\begin{takeaway}
	\textbf{Latent Bypass}: Off-the-shelf latent visual reasoning models bypass latent tokens, which have little to no causal effect on the model's final answer.
\end{takeaway}

\section{Training Data Does Not Incentivize the Use of Latent Tokens}

\label{section4}


Given the consistent results in Section \ref{section3} across models, we hypothesize that the main reason underlying the latent bypass problem comes from the choice of training data. In particular, the intermediate images tend to be subregions of the original input images (see Section~\ref{latent_comparison}), making them easily recoverable from the input image itself.
Since the oracle latents are a lossy compression of these intermediate images, we posit that they offer little incentive for the model to rely on them during training when the full information is easily recoverable from the input directly.

In this section, we first gather evidence that conditioning on latent tokens does not make the task substantially easier for the model (\textbf{Pause Tokens}), explaining why models do not rely on them. We then adjust the intermediate images in two training scenarios (\textbf{Masked Training} and \textbf{Tetris}) where we \emph{design} the latent tokens to be helpful, and demonstrate that a model trained on these datasets learns to rely on latent tokens at inference time, contrary to the previous cases. We test our hypothesis in the context of \emph{LanteRn} models \citep{viveiros2026lanternlatentvisualstructured}, since it allows encoding intermediate images independently of the original image, while also providing a simple implementation.


\label{training_data}
\textbf{Pause Tokens.} To test if oracle latent tokens based on image subregions support the model in performing the task it is trained for, we contrast \textit{LanteRn} with a baseline trained on the same data but without access to latent tokens, and setting $\gamma=0$ in Eq.~\ref{eq:latent_loss}. We use the training dataset of \textit{LanteRn} that uses subregions as intermediate images, filtering out samples solvable from text alone (see Appendix~\ref{app:lantern-filtering}), and replace oracle latent tokens with placeholder tokens during training, inspired by pause tokens \citep{goyal2024think}. This ensures that the baseline uses the same number of time steps and compute but cannot condition on the intermediate image. At inference time, we provide oracle latent tokens to \textit{LanteRn}; for the pause token baseline we use standard autoregressive decoding. Both models achieve the same accuracy (79\%, see column \textit{Std} in Table~\ref{tab:masking}) on held-out data. Hence, even under ideal in-distribution conditions with oracle latents available during evaluation, conditioning on latent tokens provides no clear benefit over the pause token baseline.
This supports our hypothesis that conditioning on intermediate images based on subregions does not significantly help the model learn the task.




\begin{figure}
	\centering
	\includegraphics[width=\linewidth]{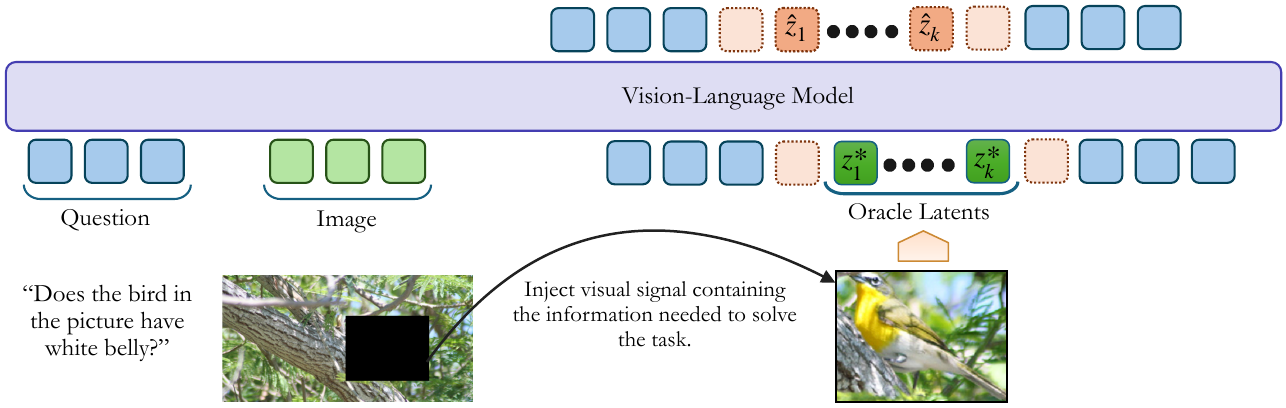}
    \caption{Setup for masked training: we mask the relevant subregion in the input image but keep the intermediate image, creating an incentive for a model to rely on latent tokens.}
	\label{fig:masking}
\end{figure}


\textbf{Masked Training.}
Next, we show that changing the training data makes a model rely on latent tokens. We start with a simplified setup, again using the \emph{LanteRn} training data. To make the intermediate images carry information beyond the input image, we mask the regions corresponding to the intermediate helper images (see Fig.~\ref{fig:masking}), maintaining the original intermediate images. 
As a result, the visual information necessary to answer the question can \emph{only} be accessed from the oracle latent tokens derived from the intermediate image, providing a clear incentive to rely on the latent tokens. 


While this setup deviates from realistic settings in that the input image alone is insufficient to solve the task, it allows us to disentangle if failures come from limitations of the modeling approach, or if the model can effectively leverage latent tokens when the intermediate image provides a strong source of task-relevant information beyond the input image.
We contrast this with the pause token baseline and the original \emph{LanteRn} setup retrained on the corresponding original, non-masked data.

As shown in Table~\ref{tab:masking}, the model trained with masked images achieves high accuracy when provided with oracle latents during inference, but accuracy drops consistently when oracle latents are replaced using the interventions from Section~\ref{section3}. This is the first setting where we observe a clear gap between oracle latents and the interventions, demonstrating that the model relies more heavily on latent tokens when the intermediate image provides additional task-relevant signal.
As expected, the pause token baseline performs relatively poorly, confirming that oracle latents carry important information.

In addition to our hypothesis about the choice of training data, these results also point to a modeling challenge:
Comparing the accuracy of \textit{LanteRn} with oracle latents across the standard and the masked training setup, we observe a drop in performance when the input image is masked. This is despite the intermediate image encoding the masked region. We attribute this drop in performance to the lossy way oracle latents are constructed (see Section~\ref{latent_comparison}).



\begin{figure}[t]
	\centering
	\begin{minipage}[b]{0.59\textwidth}
		\centering
		\includegraphics[width=\linewidth]{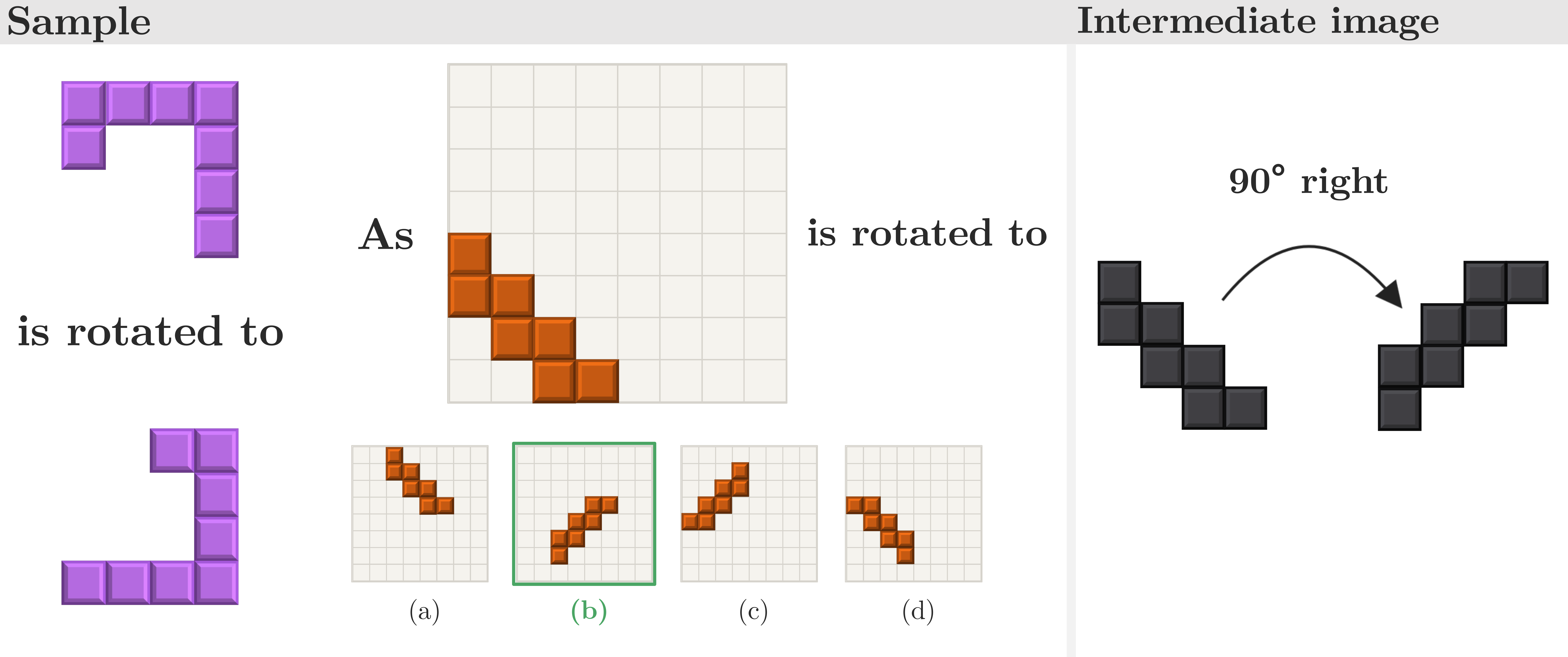}
		\caption{Sample from our Tetris-like dataset for analogical reasoning about rotations. On the right: intermediate image as source for oracle latent tokens demonstrating the rotation.}
		\label{fig:tetris_merged}
	\end{minipage}
	\hfill
	\begin{minipage}[b]{0.39\textwidth}
    \small
		\centering
		\captionof{table}{Accuracy on Tetris-like data under \textit{LanteRn} setup. 
        Instead of the subregion intervention, we use a random intermediate image.
        }
		\label{tab:tetrislan_results}
	\begin{tabular}{lc}
	\toprule
	\textbf{Latent source} & \textbf{Acc.} $\uparrow$\\
	\midrule
	Oracle latents            & \textbf{0.86}\\
	Standard Inference         & 0.33\\
	Random Inter. Image & 0.30 \\
	Noise     & 0.26\\
	Zeros                     & 0.25\\
    Random baseline                     & 0.25\\
	\midrule
	Pause Tokens                    & 0.34\\
    
	\bottomrule
\end{tabular}
		\vspace{10pt} 
	\end{minipage}
\end{figure}

\textbf{Tetris-like Rotations.}
To understand how far our observation generalizes beyond subregion regimes, we design a more realistic scenario where the intermediate image is not simply a crop of the input but provides task-relevant information that cannot be trivially recovered from the input image.
We focus on rotation as an instance of visual reasoning, and develop a synthetic dataset of transformations of \emph{Tetris-like} shapes (polyominoes).
In each sample (see Figure~\ref{fig:tetris_merged}), the model is given two polyominoes A and B (left), where B is a rotation of A. The task is to apply the rotation that maps A to B to a third polyomino C and select the correct result among four candidates. The candidates are carefully designed to include multiple plausible rotations, increasing difficulty and reducing the chance of success through superficial cues.

We generate a diverse set of shapes yielding over 8k synthetic combinations (see Appendix~\ref{app:tetris-data} for more details), from which we create 4k training and 400 evaluation instances.\footnote{All models and data are publicly available to ensure reproducibility. The models can be found at \href{https://huggingface.co/collections/AGViveiros/lantern-models}{1}, and the dataset at \href{https://huggingface.co/collections/AGViveiros/lantern-data}{2}.} To abstract away from task-irrelevant information and mirror how humans might approach this problem, we remove color from the intermediate image. We then train \emph{LanteRn} using the hyperparameters in Appendix~\ref{app:retrain-lantern}.

Table~\ref{tab:tetrislan_results} shows that the model solves the task to a reasonable extent with oracle tokens, while all interventions replacing oracle tokens with task-irrelevant information reduce accuracy to chance level (25\%, four balanced options), demonstrating that latent tokens play an important role in the model's prediction process. The pause token baseline shows a significant performance gap relative to \emph{LanteRn} with oracle tokens, confirming that the intermediate image helps in solving the task.
Overall, these results support our hypothesis that when intermediate information is not directly recoverable from the input image and instead provides a non-trivial task-relevant signal, models are more likely to rely on latent representations during inference.


\begin{takeaway}
	\textbf{Takeaway:} More challenging spatial visual tasks, such as identifying the correct rotation of an object, can benefit from providing intermediate visual steps.
\end{takeaway}

\section{Models Struggle to Predict Latent Tokens}
\label{section5}



While oracle latents lead to strong performance in the Tetris-like scenario discussed in Section \ref{section4}, the model’s generated latents during standard inference achieve performance only slightly above random chance (see Table~\ref{tab:tetrislan_results}), revealing a large gap between oracle and generated latents. This means that the model can leverage latent tokens when provided with informative representations, but additional challenges emerge when these representations must be generated at inference time.
In this section, we investigate the latent representations produced at inference time and analyze how the generated and oracle latent representations are distributed across the space.


\textbf{Methodology.} We conduct this analysis on the 300 VisCoT samples used in Section~\ref{training_data} as held-out data. For each model, we compute the corresponding oracle latent representations $\bm{z}^{*(1)}, \ldots, \bm{z}^{*(N)}$ on the held-out data using the approach of the respective framework. We then run each model with standard inference to generate latent tokens $\hat{\bm{z}}^{(i)}$ and measure their cosine similarity to all the oracle latents. For each generated latent $\hat{\bm{z}}^{(i)}$, we rank all oracle latents $\bm{z}^{*(1)}, \ldots, \bm{z}^{*(N)}$ by their cosine similarity to $\hat{\bm{z}}^{(i)}$ and compute the rate of ``retrieving'' the matching oracle latent $\bm{z}^{*(i)}$ within the top-1, top-5, and top-10 of the ranking.
\begin{wraptable}{r}{8.5cm}
	\centering
	\small
	\vspace{-4pt}
	\caption{Comparison of predicted and oracle latents. Models predict latents that are highly similar to each other (high similarity within pred) but relatively dissimilar to corresponding oracle latents, as shown by poor scores for Retrieval and USP.}
	\label{tab:retrieval_results}
	\vspace{0.cm}
	\begin{tabular}{lrrrrrr}
		\toprule
		
		\textbf{Model} & \multicolumn{3}{c}{\textbf{Retrieval \%} $\uparrow$} & \multicolumn{1}{c}{\textbf{USP} $\downarrow$} & \multicolumn{2}{c}{\textbf{Sim. Within}} \\
		\cmidrule(lr){2-4} \cmidrule(lr){5-5} \cmidrule(lr){6-7}
		& \textbf{@1}       & \textbf{@5}      & \textbf{@10}      & \textbf{\%}   & \textbf{Pred} & \textbf{Oracle} \\
		
		\midrule
        \multicolumn{7}{c}{\textit{VisCoT}} \\[0.5ex]
		ILVR & 0.3 & 3.8 & 6.9 & 99.0 & 0.82 & 0.33 \\
		LVR & \textbf{3.3} & \textbf{6.3} & \textbf{10.0} & 99.6 & 0.85 & 0.22 \\
		LanteRn SFT & 0.3 & 4.3 & 6.3 & 100.0 & 0.98 & 0.60 \\
		LanteRn RL & 0.3 & 3.0 & 5.7 & 100.0 & 0.97 & 0.60 \\
		Monet SFT & 1.0 & 2.7 & 7.0 & 100.0 & 0.91 & 0.96 \\
		Monet RL & 2.0 & 2.7 & 7.0 & 100.0 & 0.91 & 0.96 \\
		Random Baseline & 0.3 & 1.6 & 3.3 & \textbf{50.0} &-- & --\\
        \midrule
        \multicolumn{7}{c}{\textit{Tetris-like Data}} \\[0.5ex]
        LanteRn SFT & 0.0 & \textbf{1.2} & 2.2 & 100.0 & 0.98 & 0.90 \\
		Random Baseline & \textbf{0.2} & \textbf{1.2} & \textbf{2.4} & \textbf{50.0} &-- & --\\
		\bottomrule
	\end{tabular}
	\vspace{-12pt}
\end{wraptable}
To complement this, we also compute how often each predicted latent $\hat{\bm{z}}^{(i)}$ is closer to an unrelated predicted latent $\hat{\bm{z}}^{(j)}$ than to its corresponding oracle latent $\bm{z}^{*(i)}$ (unrelated self-generated preference, USP), formally defined as $\mathbb{E}_{i, j \neq i} \mathbb{1}(\text{sim}(\hat{\bm{z}}^{(i)}, \hat{\bm{z}}^{(j)}) > \text{sim}(\hat{\bm{z}}^{(i)}, \bm{z}^{*(i)}) )$, where $\mathbb{1}(\cdot)$ is the indicator function.
Finally, to get an overview of the overall layout of latent representations, we measure the average pairwise similarity within the set of predicted latent tokens, i.e. $\mathbb{E}_{i, j \neq i}\text{sim}(\hat{\bm{z}}^{(i)}, \hat{\bm{z}}^{(j)})$ and contrast this with an analogous metric for the oracle latent tokens, i.e. $\mathbb{E}_{i, j \neq i}\text{sim}(\bm{z}^{*(i)}, \bm{z}^{*(j)})$.



\textbf{Results.} The results in Table \ref{tab:retrieval_results} are consistent across models and reveal that predicted latents rarely align with their corresponding oracle representations, with retrieval accuracy of 10\% or lower even in the top-10. USP is close to 100\% for all models, meaning predicted latents are almost always closer to other predicted latents than to the ground-truth.
Predicted latents are all highly similar to one another (similarity between 0.8 and 0.98), while oracle latents tend to be more diverse, indicating that latent predictions collapse to a narrow region of space rather than capturing the oracle's variety. Monet is an exception, with high similarity in both predicted and oracle latents. On the Tetris-like data, where the retrained LanteRn model relies on latent tokens, the picture is similar to VisCoT, except that oracle latents are more similar to each other on average because the intermediate images occupy a narrower region of space of the vision encoder.

\begin{figure}
	\centering
	\includegraphics[width=0.49\linewidth]{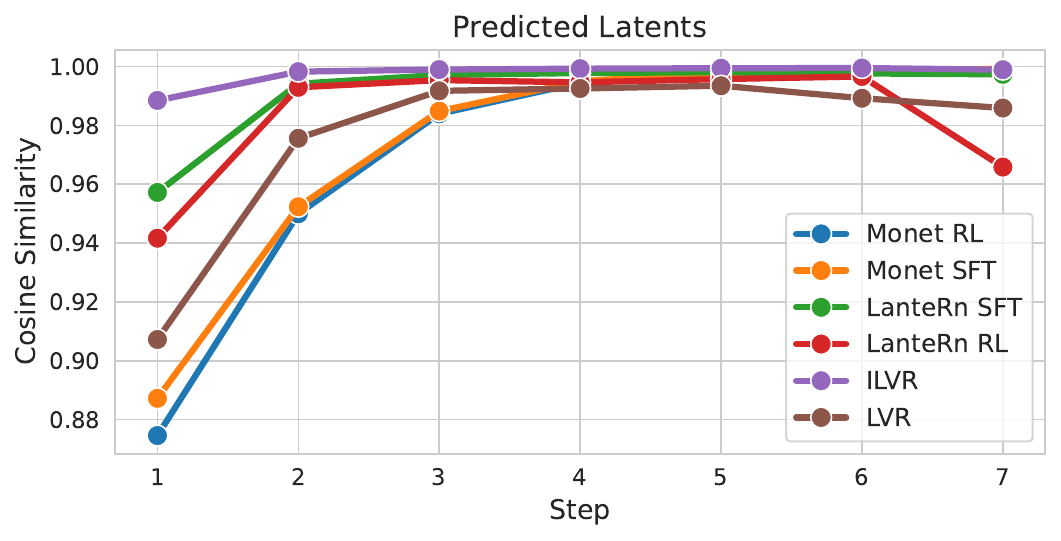}
	\includegraphics[width=0.49\linewidth]{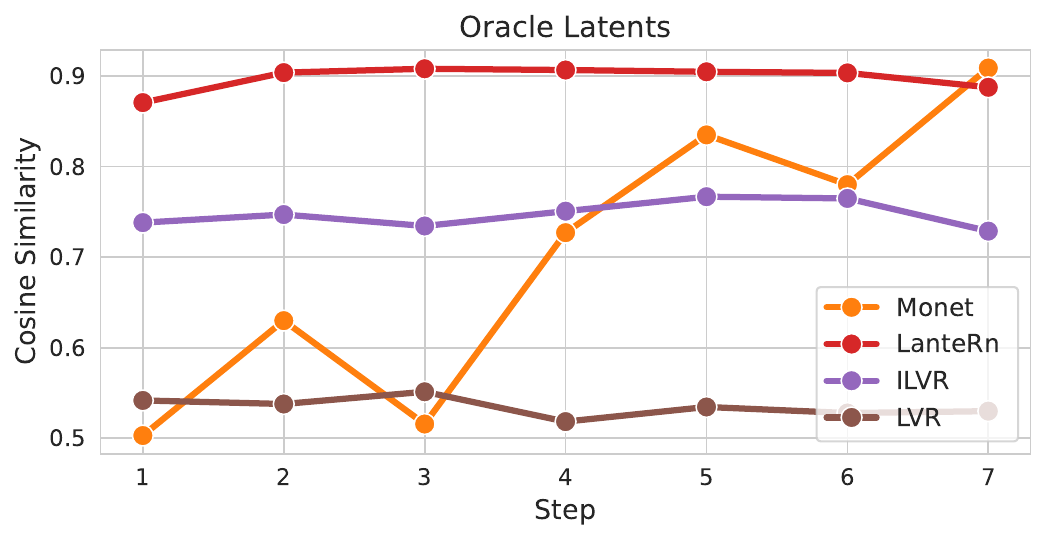}
	\caption{Average cosine similarity between consecutive time steps of latent representations for predicted latents (left, $\text{sim}(\hat{\bm{z}}_{t}, \hat{\bm{z}}_{t+1})$) and oracle latents (right, $\text{sim}(\bm{z}^*_{t}, \bm{z}^*_{t+1})$). Predicted latents collapse to highly similar representations with time.
    }
	\label{fig:timesteps}
\end{figure}

In Figure~\ref{fig:timesteps}, we show cosine similarity between consecutive latent positions. For all models, predicted latents collapse to increasingly similar representations over time, a trend not reflected in the oracle latents, which maintain lower and stable inter-step similarity. The exception is Monet, where oracle latents also grow more similar over time, possibly because Monet is the only approach that generates oracle latents autoregressively from a teacher model. Experiments with partial teacher forcing suggest that the growing collapse of predicted latents is partly caused by errors accumulating during autoregressive decoding (Appendix~\ref{app:teacher-forcing}).




\begin{takeaway}
	\textbf{Latent Representation Collapse:} Predicted latent tokens collapse to highly similar representations that are relatively dissimilar to their corresponding oracle tokens.
\end{takeaway}

\section{Discussion and Limitations}
\label{sec:discussion}

Our findings point to two challenges for future work on latent visual reasoning: How to make oracle latent tokens more useful for prediction to avoid the latent bypass problem (raised in Section~\ref{section3}), and how to improve the prediction of latent tokens to overcome the latent token collapse (Section~\ref{section5}). 

To avoid the latent bypass problem, our findings with the dataset of Tetris-like rotations point to the creation of training datasets more suitable for latent visual reasoning. In designing such datasets, intermediate steps need to provide a helpful scaffolding for models that are intended as the basis for latent visual reasoning. This can be tested during dataset creation with simple baselines, such as pause tokens, that do not condition on the intermediate images. Promising candidates are tasks that require imagining or transforming visual content, such as mental rotation and folding, visual search in puzzles, or predicting future states in navigation, where latent tokens encode task-relevant visual operations rather than information already visible in the input.
Models that learn to extract more information from latent tokens during training on such datasets will increasingly depend on high-quality latent tokens at inference time. As a simple method to diagnose latent representation collapse, we recommend monitoring the similarity within the predicted latents and between predicted and oracle latents (see Section~\ref{section5}).


While our analysis shows very consistent results across four approaches with models in two scales (3B and 7B) and different training stages (SFT and RL), it remains an open question to what extent substantially different design decisions in the supervision signal or the base model can affect when models bypass latent tokens. Given the rapidly growing field, a comprehensive analysis of all existing models is infeasible; instead, we encourage future work to test if models bypass latent tokens as part of the evaluation.
Finally, our experiments with the Tetris-like data suggest geometric transformations as a plausible domain for latent visual reasoning but what a comprehensive real-world dataset for kickstarting latent visual reasoning could look like remains an open problem for future work. 

\section{Related Work}
Latent Visual Reasoning has become a very active area of research. In addition to the methods we reviewed in Section~\ref{latent_comparison}, other methods of note are \textit{Mirage} \citep{yang2025machine}, \textit{IVT-RL} \citep{chen2025reasoning}, \textit{Laser} \citep{wang2026forest} and \textit{VaLR} \citep{jeon2026vision}.
Several works have investigated which role latent tokens play in latent reasoning, particularly for text-only models.
\citet{zhang2025latent} consider \textit{Coconut} \citep{hao2024training}, a text-only approach, and demonstrate that it is more prone to learning shortcuts than supervised fine-tuning with CoTs. Their findings also suggest that latent reasoning tokens have limited causal influence on the model's final answer, at least when the model is trained for logical reasoning and tested on questions probing for world knowledge.
\citet{dilgren2026are} perform a similar analysis, finding that latent token sequences can be shortened to a large extent or removed entirely without impacting performance on logical reasoning datasets. However, their results suggest that latent tokens play a larger role in mathematical reasoning.
In contrast, our work investigates the impact of latent tokens of latent \emph{visual} reasoning models and \textit{why} models do not rely on them.

In contemporary work, \citet{li2026imaginationhelpsvisualreasoning} also find that latent visual tokens have limited causal impact on the decision process of two latent visual reasoning models (\textit{Monet}, \textit{Mirage}). \citet{li2026imaginationhelpsvisualreasoning} then analyze the latent tokens predicted by \textit{LVR}, \textit{Mirage} and \textit{Monet} at inference time.
They show that latents have very high similarity to each other both across instances (resembling our similarity metric within predicted latents in Table~\ref{tab:retrieval_results}) and within the latent reasoning chain (similar to Figure~\ref{fig:timesteps} left).
However, their analysis neither considers the relation between predicted and oracle latents nor the geometry of the oracle latents. Moreover, they focus on models at inference time and their findings are consistent with the hypothesis that the bottleneck in latent reasoning is the difficulty to predict high-quality latents. 
In this work, we show that the issue is more fundamental and that four recent models (\textit{LVR}, \textit{Monet}, \textit{LanteRn}, \textit{ILVR}) also ignore \emph{oracle} latent tokens and we investigate \emph{why} that is the case. Our analysis shows that a major bottleneck lies in the training data. In fine-tuning experiments on diagnostic datasets, we find that the latent tokens do play an important role when the oracle latent tokens contain information that sufficiently supports the reasoning process.

\section{Conclusion}
We investigate the causal role of latent tokens in four recent methods for latent visual reasoning. Surprisingly, we find that models largely do not take latent tokens into account for their answers. We then analyze why this is the case and identify that oracle latent tokens used in common setups provide little support for models to predict the answer during training, leading to models bypassing the latents and ignoring them. We construct a diagnostic fine-tuning dataset and show that latent tokens play a significant role in a model trained in this context. Finally, we find that latents produced by these models collapse to highly similar representations that are relatively dissimilar to their corresponding oracle latents.

\section*{Acknowledgments}
This work was supported by the project DECOLLAGE (ERC-2022-CoG 101088763) and by FCT/MECI through national funds and when applicable co-funded EU funds under UID/50008: Instituto de Telecomunicações. We thank EuroHPC for the HPC resources used to support this work through grants EHPC-AI-2024A05-044 and E-EXT-2025E01-100.

\bibliographystyle{plainnat}  
\bibliography{references}
\newpage


\appendix

\section{Additional Results}
\label{app:additional-results}
\begin{table}[h]
    \centering
        \caption{Performance on Blink and V$^*$ subsets in tabular format, supplementing Figure ~\ref{fig:interventions}.}
    \label{tab:placeholder}
\begin{tabular}{llrrrrrr}
\toprule
 & & ILVR 7B & LVR 7B & \multicolumn{2}{r}{LanteRn 3B} & \multicolumn{2}{r}{Monet 7B} \\
 & & SFT & RL & RL & SFT & RL & SFT \\
 & Intervention &  &  &  &  &  &  \\
\midrule
\multirow[t]{5}{*}{Blink} & Noise & 0.62 & 0.75 & 0.67 & 0.65 & 0.61 & 0.57 \\
 & Random Subregion & 0.59 & 0.70 & 0.67 & 0.65 & 0.61 & 0.59 \\
 & Skip Latents & 0.60 & 0.73 & 0.65 & 0.65 & 0.60 & 0.57 \\
 & Standard Inference & 0.63 & 0.68 & 0.65 & 0.65 & 0.63 & 0.60 \\
 & Zeros & 0.61 & 0.76 & 0.66 & 0.62 & 0.64 & 0.50 \\
\cline{1-8}
\multirow[t]{5}{*}{V*} & Noise & 0.60 & 0.81 & 0.70 & 0.64 & 0.80 & 0.76 \\
 & Random Subregion & 0.61 & 0.81 & 0.71 & 0.63 & 0.79 & 0.77 \\
 & Skip Latents & 0.46 & 0.81 & 0.71 & 0.62 & 0.78 & 0.78 \\
 & Standard Inference & 0.63 & 0.80 & 0.72 & 0.63 & 0.82 & 0.75 \\
 & Zeros & 0.65 & 0.80 & 0.70 & 0.60 & 0.80 & 0.77 \\
\cline{1-8}
\bottomrule
\end{tabular}

\end{table}

\section{Statistical Significance of the Interventions}
\label{app:significance}
Our interventions are paired by construction: each condition is evaluated on the same items with the same model, and only differ how/if we intervene on the latent tokens. For each model, benchmark and intervention, we therefore run a one-sided exact binomial test on the paired predictions, with the alternative hypothesis that standard inference outperforms the intervention. For a model to demonstrably rely on its own latents, standard inference would have to significantly outperform \emph{every} intervention. We therefore combine the tests with an intersection-union test and report the maximum $p$-value across interventions (Table~\ref{tab:significance}). None of the 18 model--benchmark pairs is significant at $\alpha=0.05$.

Individual interventions sometimes reduce accuracy significantly (e.g., ILVR on VisCoT: Zeros, $-2.9$pp, $p<.001$; Skip Latents, $-2.2$pp, $p<.001$), but these effects are not consistent: for every model, some intervention performs as well as or nominally better than standard inference (e.g., ILVR on VisCoT, Random Subregion: $+1.3$pp, $p=.955$).

\begin{table}[h]
\centering
\small
\caption{Maximum one-sided $p$-value across interventions (intersection-union test, $\alpha=0.05$). For no model--benchmark pair does standard inference significantly outperform all interventions.}
\label{tab:significance}
\begin{tabular}{lrrr}
\toprule
\textbf{Model} & \textbf{VisCoT} ($N{=}3000$) & \textbf{BLINK} ($N{=}265$) & \textbf{$V^*$Bench} ($N{=}191$) \\
\midrule
ILVR-7B     & .955  & .411  & .750 \\
LVR-7B      & 1.000 & 1.000 & .746 \\
Monet SFT   & 1.000 & .423  & .820 \\
Monet RL    & .977  & .774  & .500 \\
LanteRn SFT & .930  & .432  & .806 \\
LanteRn RL  & .982  & .981  & .500 \\
\bottomrule
\end{tabular}
\end{table}

\section{Additional Benchmarks: GQA and BabyVision}
\label{app:more-benchmarks}
To test whether latent bypass also occurs on tasks requiring more explicit multi-step reasoning, we evaluate on two additional benchmarks. GQA \citep{hudson2019gqa} is a compositional VQA benchmark whose questions are associated with multi-step reasoning programs, and it is close to the training distribution of the evaluated models. BabyVision \citep{chen2026babyvision} tests visual reasoning and manipulation in a substantially different, synthetic domain; we use its multiple-choice questions. Table~\ref{tab:more-benchmarks} shows results and also reports how often each model produces any latent tokens at all (\emph{Lat.\,\%}).

On GQA, every model has at least one intervention that performs about as well as or better than standard inference. Monet SFT degrades strongly under Noise, Zeros and Skip Latents, but stays within 0.4pp under Random Subregion. ILVR produces latent tokens for only 1.1\% of samples and thus answers almost entirely through the text pathway, suggesting that GQA is out-of-distribution for ILVR.

BabyVision is out-of-distribution for all models, which often skip the latent pathway, making it a less clean test of latent usage. Still, the picture is similar: for every model except Monet, at least one intervention performs at least as well as standard inference. Monet SFT and, to a lesser extent, Monet RL do better with standard inference, but both remain below the majority baseline of always predicting the most frequent option (38.5\%).

\begin{table}[h]
\centering
\small
\setlength{\tabcolsep}{4pt}
\caption{Accuracy (\%) under interventions on GQA and BabyVision; best accuracy per model in bold. \emph{Lat.\,\%}: fraction of samples in which the model produces latent tokens under standard inference.}
\label{tab:more-benchmarks}
\begin{tabular}{llrrrrrr}
\toprule
& \textbf{Model} & \textbf{Standard} & \textbf{Rand.\ Subregion} & \textbf{Noise} & \textbf{Zeros} & \textbf{Skip Latents} & \textbf{Lat.\,\%} \\
\midrule
\multirow{6}{*}{GQA}
& LanteRn SFT & 79.2 & 78.8 & 79.0 & \textbf{79.4} & 78.6 & 100.0 \\
& LanteRn RL  & 81.2 & \textbf{81.9} & 81.6 & 81.8 & 81.8 & 100.0 \\
& LVR-7B      & 78.1 & 76.7 & 81.1 & 80.5 & \textbf{83.7} & 100.0 \\
& Monet SFT   & \textbf{70.2} & 69.8 & 41.7 & 33.8 & 33.1 & 97.0 \\
& Monet RL    & 74.7 & 74.5 & 75.0 & \textbf{75.5} & 75.1 & 97.0 \\
& ILVR        & \textbf{83.9} & \textbf{83.9} & \textbf{83.9} & \textbf{83.9} & \textbf{83.9} & 1.1 \\
\midrule
\multirow{6}{*}{BabyVision}
& LanteRn SFT & 20.7 & 20.7 & 20.7 & 20.7 & \textbf{21.5} & 31.9 \\
& LanteRn RL  & 21.5 & 20.0 & 21.5 & 21.5 & \textbf{22.2} & 100.0 \\
& LVR-7B      & 20.0 & 30.4 & \textbf{34.8} & 34.1 & 31.1 & 100.0 \\
& Monet SFT   & \textbf{36.3} & 31.1 & 29.1 & 29.6 & 30.4 & 82.6 \\
& Monet RL    & \textbf{34.8} & 28.9 & 33.3 & 30.4 & 33.3 & 57.8 \\
& ILVR        & 34.8 & 31.9 & 37.5 & \textbf{43.7} & 34.8 & 62.6 \\
\bottomrule
\end{tabular}
\end{table}

\section{Attention to Latent Positions}
\label{app:attention}
To see whether models attend to latent positions at all, we measure the fraction of attention that (i) the reasoning tokens generated right after the latent block and (ii) the final answer tokens allocate to latent positions, averaged over heads and aggregated into layer quartiles (Table~\ref{tab:attention}). We report LanteRn SFT, LVR-7B and Monet RL as representatives of the different training paradigms. All three allocate a small but non-zero share of attention to latent positions, with similar patterns for reasoning and answer tokens. Attention weights alone, however, do not show whether the attended representations contribute useful information; this analysis therefore complements the interventions in Section~\ref{section3}, which test causal reliance directly.

\begin{table}[h]
\centering
\small
\setlength{\tabcolsep}{4pt}
\caption{Attention (\%) allocated to latent positions by post-latent reasoning tokens and by answer tokens, per layer quartile. \emph{Own}: self-generated latents; \emph{Oracle}: oracle latents; \emph{Zeros}: zero intervention.}
\label{tab:attention}
\begin{tabular}{llrrrrrcrrrrr}
\toprule
& & \multicolumn{5}{c}{\textbf{Post-latent reasoning tokens}} & & \multicolumn{5}{c}{\textbf{Answer tokens}} \\
\cmidrule(lr){3-7} \cmidrule(lr){9-13}
\textbf{Model} & \textbf{Latents} & Q1 & Q2 & Q3 & Q4 & Mean & & Q1 & Q2 & Q3 & Q4 & Mean \\
\midrule
\multirow{3}{*}{LanteRn SFT}
& Own    & 0.05 & 0.44 & 0.86 & 0.49 & 0.46 & & 0.02 & 0.14 & 0.40 & 0.43 & 0.24 \\
& Oracle & 0.10 & 0.52 & 1.37 & 0.68 & 0.67 & & 0.02 & 0.18 & 0.60 & 0.55 & 0.34 \\
& Zeros  & 4.34 & 1.95 & 1.48 & 1.16 & 2.23 & & 2.52 & 1.06 & 0.93 & 1.23 & 1.43 \\
\midrule
\multirow{3}{*}{LVR-7B}
& Own    & 3.17 & 3.45 & 6.65 & 2.81 & 4.02 & & 1.38 & 1.67 & 2.72 & 1.60 & 1.84 \\
& Oracle & 3.43 & 3.49 & 6.52 & 2.46 & 3.97 & & 1.54 & 1.79 & 3.35 & 1.44 & 2.03 \\
& Zeros  & 5.44 & 4.30 & 7.60 & 3.46 & 5.20 & & 3.01 & 2.49 & 3.91 & 1.97 & 2.84 \\
\midrule
\multirow{3}{*}{Monet RL}
& Own    & 2.55 & 2.02 & 1.81 & 1.29 & 1.92 & & 1.08 & 1.44 & 1.25 & 0.81 & 1.14 \\
& Oracle & 0.88 & 1.92 & 2.47 & 1.89 & 1.79 & & 0.30 & 1.10 & 0.99 & 0.90 & 0.82 \\
& Zeros  & 8.67 & 6.10 & 5.95 & 7.61 & 7.08 & & 2.12 & 0.92 & 1.13 & 2.16 & 1.58 \\
\bottomrule
\end{tabular}
\end{table}

\section{Error Accumulation in Latent Prediction}
\label{app:teacher-forcing}
To test whether the growing similarity between consecutive predicted latents (Figure~\ref{fig:timesteps}) stems from error accumulation, we run LVR with partial teacher forcing: for the first $k$ steps we feed oracle latents, and from step $k{+}1$ onward the model's own predictions. Table~\ref{tab:teacher-forcing} reports the cosine similarity between each predicted latent and the oracle latent it should match; $k=0$ is fully autoregressive generation. Two trends point to error accumulation: down each column, a longer oracle prefix yields predictions closer to the oracle, and along each row, similarity tends to drop once autoregressive decoding begins (position $>k$).

\begin{table}[h]
\centering
\small
\setlength\tabcolsep{3pt}
\renewcommand\arraystretch{1.15}
\caption{Cosine similarity between predicted and oracle latents for LVR under partial teacher forcing with an oracle prefix of length $k$. Values at positions $\leq k$ repeat across rows because the model receives the same oracle prefix.}
\label{tab:teacher-forcing}
\begin{tabular}{c|*{7}{c}}
\toprule
Oracle & \multicolumn{7}{c}{\textsc{Latent position}} \\
prefix $k$ & 1 & 2 & 3 & 4 & 5 & 6 & 7 \\
\midrule
0 & \bc{.484} & \bc{.495} & \bc{.482} & \bc{.488} & \bc{.469} & \bc{.449} & \bc{.430} \\
1 & \bc{.484} & \bc{.507} & \bc{.508} & \bc{.514} & \bc{.494} & \bc{.471} & \bc{.451} \\
2 & \bc{.484} & \bc{.507} & \bc{.522} & \bc{.530} & \bc{.499} & \bc{.474} & \bc{.468} \\
3 & \bc{.484} & \bc{.507} & \bc{.522} & \bc{.539} & \bc{.506} & \bc{.480} & \bc{.473} \\
4 & \bc{.484} & \bc{.507} & \bc{.522} & \bc{.539} & \bc{.520} & \bc{.489} & \bc{.473} \\
5 & \bc{.484} & \bc{.507} & \bc{.522} & \bc{.539} & \bc{.520} & \bc{.493} & \bc{.479} \\
6 & \bc{.484} & \bc{.507} & \bc{.522} & \bc{.539} & \bc{.520} & \bc{.493} & \bc{.478} \\
\bottomrule
\end{tabular}
\end{table}

\section{Model Checkpoints}
\label{app:model-ckpts}
We evaluate the following models in our experiments:

\begin{table}[h]
\tiny
\centering
\renewcommand{\arraystretch}{1.2}
\begin{tabular}{llll}
\hline
\textbf{Model} & \textbf{Size} & \textbf{Training / Variant} & \textbf{Link} \\
\hline
LVR-7B & 7B & Base & \url{https://huggingface.co/vincentleebang/LVR-7B} \\
Monet-RL-7B & 7B & RL & \url{https://huggingface.co/NOVAglow646/Monet-7B} \\
Monet-SFT-7B & 7B & SFT (Stage 2) & \url{https://huggingface.co/NOVAglow646/Monet-SFT-7B/tree/main/stage2} \\
Monet-SFT-7B & 7B & SFT (Stage 3) & \url{https://huggingface.co/NOVAglow646/Monet-SFT-7B/tree/main/stage3} \\
ILVR & 7B & SFT-Stage2 & \url{https://huggingface.co/shuai22/comt_ckpt} \\
LanteRn-3B-SFT & 3B & SFT & \url{https://huggingface.co/AGViveiros/LanteRn-3B-SFT} \\
LanteRn-3B-RL & 3B & RL & \url{https://huggingface.co/AGViveiros/LanteRn-3B-RL} \\
\hline
\end{tabular}
\caption{Models evaluated in this work, grouped by architecture family and training stage.}
\label{tab:models}
\end{table}

\section{Details on Extraction of Random Subregion/Intermediate Images}
\label{app:subregion-extraction}
For the Random Subregion intervention, when intervening on a sample $s_t$, we use the subsequent sample $s_{t+1}$ and treat its intermediate image as the ground-truth intermediate for $s_t$ whenever it is available. If the intermediate image is not available, we instead extract a subregion from the input image $s_{t+1}$ and use it as a proxy for the intermediate representation.

Since most of the evaluated models were trained on the Viscot dataset, we further ensure that the extracted subregions approximately match the size distribution and aspect ratios of intermediate images in that dataset, in order to better align the intervention with the training data statistics. After extracting each intermediate image, we process it through the corresponding training framework to obtain the associated oracle latent tokens.

\section{Details on Extraction of Oracle Latents}
\label{app:oracle-extraction}
To compute oracle latents for Monet, we use latent representations precomputed from the Monet Stage-2 model \emph{(Monet-SFT-7B/stage2)}. For each sample, we run a forward pass with \textit{latent\_mode=True} and \textit{output\_hidden\_states=True}, allowing the model to produce hidden states at every layer. The full sequence is processed, including the auxiliary crop image placed within the latent block, under a specific attention mask that enforces the correct visibility structure between visual, latent, and text tokens.

The model returns hidden states for all transformer layers at the latent positions. From these, we extract only the last-layer hidden states, which are used as the oracle latent representation (shape: (latent\_size, H)). These representations are saved per sample as .pt files and later injected during evaluation as gt\_latent\_embeds, fully bypassing the model’s own latent generation.

We additionally experimented with alternative layers, including intermediate layers and averaged combinations across layers. These variations yielded no meaningful differences in performance, so we adopt the last-layer representation for consistency with the other models evaluated in this work.

\section{Details on Filtering LanteRn Dataset}
\label{app:lantern-filtering}
We start from the original LanteRn dataset, that contains $143,024$ samples and remove all instances that can be solved using the question alone, reducing the dataset by approximately 30\%. Specifically, we use \texttt{Qwen/Qwen3-VL-235B-A22B-Instruct-FP8}, prompting it to answer each question without access to the original image, discarding all samples it answers correctly.
From the dataset, we use a held-out set of 5k samples for evaluation.

\section{Details for Retraining LanteRn with Tetris-Like data}
\label{app:retrain-lantern}

Training is performed on a single node with 4 $\times$ NVIDIA GH200 GPUs (98GB memory each). 
We train for 15 epochs. We summarize the essential hyperparameters used for retraining LanteRn with Tetris-like data. 

\begin{table}[h]

\centering

\small

\begin{tabular}{ll}

\hline

\textbf{Category} & \textbf{Parameter} \\

\hline

\multicolumn{2}{l}{\textbf{Model}} \\

model\_id & Qwen2.5-VL-3B \\

latent\_size & 8 \\

\\

\multicolumn{2}{l}{\textbf{Training}} \\

num\_train\_epochs & 15 \\

learning\_rate & 1e-5 \\

lr\_scheduler\_type & cosine \\

warmup\_ratio & 0.05 \\

gamma & 0.05 (default: 0.1) \\

latent\_loss\_type & mse \\

bf16 & True \\

fp16 & False \\

freeze\_vision\_tower & True \\

freeze\_merger & True \\

freeze\_llm & False \\

eval\_steps & 80 \\

per\_device\_train\_batch\_size & 6 \\

gradient\_accumulation\_steps & 8 \\

seed & 42 \\

dataloader\_num\_workers & 4 \\

\\

max\_train\_samples & 4000 \\

use\_lvr & True \\

\hline

\end{tabular}

\caption{Hyperparameters for retraining LanteRn with Tetris-like data.}
\label{tab:lantern_tetris_params}

\end{table}

\section{Details on Tetris-like data}
\label{app:tetris-data}

\begin{figure}
    \centering
    \includegraphics[width=\linewidth]{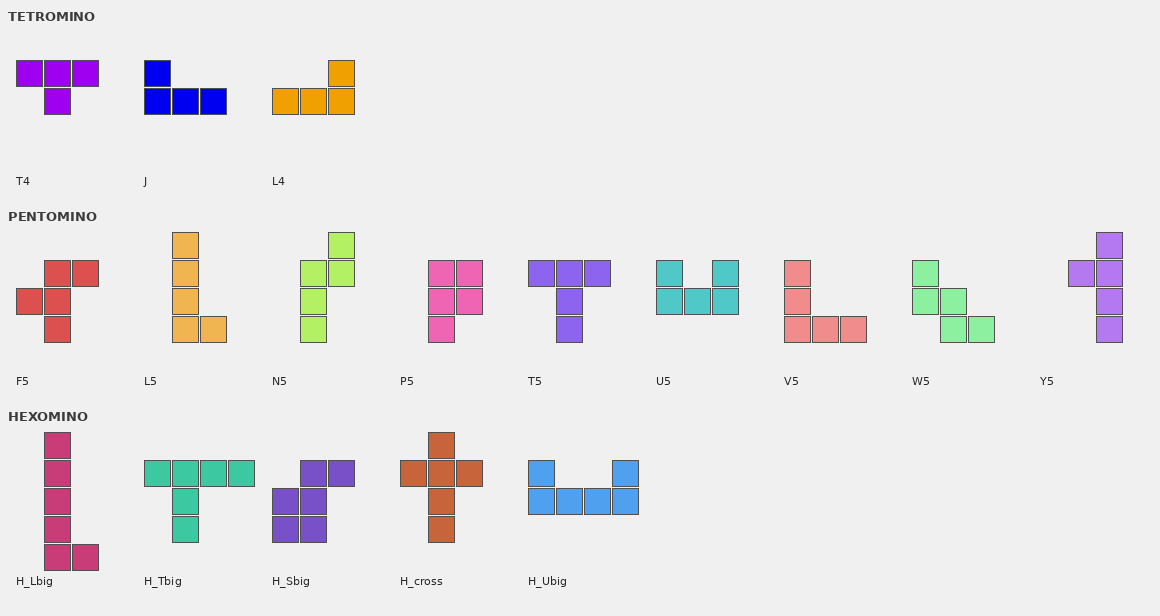}
    \caption{Possible figure combinations in the Tetris-like dataset}
    \label{fig:tetris_shapes}
\end{figure}

We develop a synthetic dataset of 8K unique samples drawn from a diverse pool of object shapes (see Figure \ref{fig:tetris_shapes}). 
We use 4,000 training samples and 400 evaluation samples.
The objects are distributed across three shape families: pentominoes ($\sim$41\% of the dataset), hexominoes ($\sim$35\%), and tetrominoes ($\sim$24\%).\
Each sample contains four candidate options, exactly one of which is correct. The labels are perfectly balanced across options (a/b/c/d $\approx$ 25\% each).

Each sample follows the format below:
\begin{verbatim}
{
  "question": "Image (A) is to image (B) as image (C) 
  is to which of the following options?
  The transformation from (A) to (B) is: 270° clockwise rotation.
    Options: (a) Option a
             (b) Option b
             (c) Option c
             (d) Option d",
  "answer": "a",
  "dataset": "tetris_analogy",
  "transform_type": "rotation",
  "transform_description": "270° clockwise rotation",
  "shape_A_name": "H_Tbig",
  "shape_C_name": "Y5",
  "shape_A_family": "hexomino",
  "shape_C_family": "pentomino",
  "option_transforms": {
    "a": "270° clockwise rotation",
    "b": "90° clockwise rotation",
    "c": "180° clockwise rotation",
    "d": "identity"
  },
  "intermediate_key": "Y5_270"
}
\end{verbatim}

\section{Evaluation Protocol for Monet-RL}
\label{app:evaluation_protocol_monet_rl}

We evaluate Monet-RL on the VisCoT, BLINK, and VStar benchmarks. For VisCoT, we use the LanteRn multiple-choice held-out set, while BLINK and VStar are evaluated on their standard test splits in the same multiple-choice format.

Since Monet-RL is trained to emit boxed\{answer\} as part of its post-latent reasoning chain, it does not reliably produce this token when the latent block is perturbed or removed, e.g., under \emph{Skip\_Latents}, \emph{Zeros}, or \emph{Random}. In such cases, the model may terminate after its visual observation without yielding an extractable answer.

\paragraph{Example (BLINK).}
\begin{quote}
\ttfamily
<|im\_start|> To get a clearer view of the sandwich and the bounding boxes, I will generate a zoomed-in image of the sandwich area. <latent\_start><latent\_end> sandwich area zoomed in for better analysis, where bounding box A is highlighted. <|im\_end|>
\end{quote}

To ensure fair and consistent answer extraction across all benchmarks and interventions, we apply a forced completion step whenever boxed\{answer\} is missing from the model output. 
Specifically, we append the suffix

\begin{quote}
\ttfamily
Therefore, the final answer is boxed\{
\end{quote}
and then decode greedily until the EOS. The resulting output becomes:

\begin{quote}
\ttfamily
... Bounding box A is highlighted. Therefore, the final answer is boxed\{A\}. <|im\_end|>
\end{quote}

This procedure is applied exclusively to \emph{Monet-RL} and is unnecessary for SFT-stage models, which produce boxed\{\} consistently regardless of the latent intervention.

\section{Technical Details of Section \ref{section5}}
\label{app:technical-details-section5}
For all results in Section~\ref{section5}, we first truncate predicted and oracle tokens to the same number of time steps where necessary because some models can generate a variable number of latent tokens at inference time (ILVR). For the results in Table~\ref{tab:retrieval_results}, we flatten the resulting representations before computing cosine similarity. There is no need for flattening for results in Figure~\ref{fig:timesteps} because we compare individual time steps.



\end{document}